\documentclass{article} 
\usepackage{iclr2027_conference,times}

\usepackage{amsmath,amsfonts,bm}

\def\eqref#1{equation~\ref{#1}}

\def\1{\bm{1}}

\DeclareMathAlphabet{\mathsfit}{\encodingdefault}{\sfdefault}{m}{sl}
\SetMathAlphabet{\mathsfit}{bold}{\encodingdefault}{\sfdefault}{bx}{n}

\usepackage{hyperref}
\usepackage{url}
\usepackage{graphicx}
\usepackage{algorithm}
\usepackage{algpseudocode}
\usepackage{amsmath,amssymb}
\algnewcommand{\Phase}[1]{\Statex \textbf{// #1}}

\usepackage{booktabs}
\usepackage{multirow}
\usepackage{pifont}
\usepackage{xspace}
\usepackage{bbm}
\usepackage{wrapfig}
\usepackage{subcaption}

\usepackage{tikz}
\usetikzlibrary{arrows.meta, positioning}

\usepackage[most]{tcolorbox}
\usepackage{listings}

\lstdefinestyle{promptstyle}{
    basicstyle=\ttfamily\scriptsize,
    breaklines=true,
    breakatwhitespace=false,
    columns=fullflexible,
    keepspaces=true,
    showstringspaces=false,
    frame=none,
    aboveskip=0pt,
    belowskip=0pt
}

\usepackage[table]{xcolor}
\definecolor{light-gray}{gray}{0.90}
\usepackage{listings}
\usepackage{tcolorbox}
\usepackage{enumitem}

\definecolor{poolblue}{HTML}{1A73E8}
\definecolor{auditred}{HTML}{D32F2F}
\definecolor{ansgreen}{HTML}{2E7D32}
\definecolor{dropgray}{HTML}{757575}
\definecolor{graybox}{HTML}{F5F5F5}

\tcbuselibrary{listings,breakable}

\newtcolorbox{querybox}{colback=graybox,colframe=black!30,fonttitle=\bfseries,title=User Query,boxrule=0.5pt}
\newtcolorbox{poolbox}[1][]{colback=blue!3,colframe=poolblue,fonttitle=\bfseries,title=#1,boxrule=0.8pt}
\newtcolorbox{auditbox}{colback=red!3,colframe=auditred,fonttitle=\bfseries,title=Self-Reflection,boxrule=0.8pt}
\newtcolorbox{sqlbox}[1][]{colback=graybox,colframe=black!40,fonttitle=\bfseries\ttfamily,title=#1,boxrule=0.5pt}

\newcommand{\model}{\textsc{PlanPool}\xspace}

\title{Beyond Correctness: Resolving Underspecification in Agentic Text-to-SQL}

\author{Wen-Zhi Li \thanks{Work done during an internship at Amazon Web Services.} \\
Department of Computer Science\\
Cornell University\\
\texttt{wenzhi@cs.cornell.edu} \\
\And
Yue Gong, Konstantinos Kanellis, \\
\textbf{Balakrishnan Murali Narayanaswamy} \\
Amazon Web Services \\
\texttt{\{yuegongy,kkanelli,muralibn\}@amazon.com}
}

\iclrfinalcopy 
\begin{document}

\maketitle

\begin{abstract}
Agentic Text-to-SQL systems can interact with users to clarify underspecified queries before generating SQL. However, a correct execution result does not necessarily imply that the agent has adequately resolved the underlying underspecification: the agent may silently make unverified assumptions that happen to match the intended answer. We show that this behavior is driven in part by premature clarification termination. Although forcing an agent to ask more questions improves execution accuracy, ambiguities are concentrated in earlier interactions, making brute-force questioning inefficient. More importantly, even when explicitly prompted to plan its clarification process, the agent frequently abandons questions that it has already identified as relevant. To address this failure mode, we introduce \model, which externalizes the clarification plan as a mutable question pool. Every planned question must be explicitly asked or dropped before submission, while newly discovered ambiguities can be added during interaction. Across three benchmarks derived from BIRD-Interact and Spider, \model consistently improves ambiguity coverage and reduces silent failures over unconstrained and prompt-based alternatives, while maintaining competitive execution accuracy. Our results highlight an important distinction in agentic reasoning: identifying missing information is not sufficient, and the agent must also reliably maintain and resolve it before committing to an answer.
\end{abstract}

\section{Introduction}

Text-to-SQL systems translate natural-language user requests into executable SQL queries over structured databases. Most existing systems assume a single-turn setting: the user provides a sufficiently specified request, and the system directly generates a SQL query~\cite{spider, bird}. In practice, however, user requests are often underspecified~\cite{ambiqt,ambrosia,practiq}. 
A request to ``rank the products” may omit the ranking criterion or direction; and domain-specific concepts may require information that is not available from the database schema alone. 
In such cases, generating SQL is inherently interactive: a reliable system must recognize what information is missing, decide whether it should be obtained from the database or from the user, and proactively ask clarification questions before committing to an implementation~\cite{chen2025learning,ambisql}.

\begin{figure}[!t]
    \centering

    \begin{subfigure}[t]{0.4\textwidth}
        \centering
        \includegraphics[width=\linewidth]{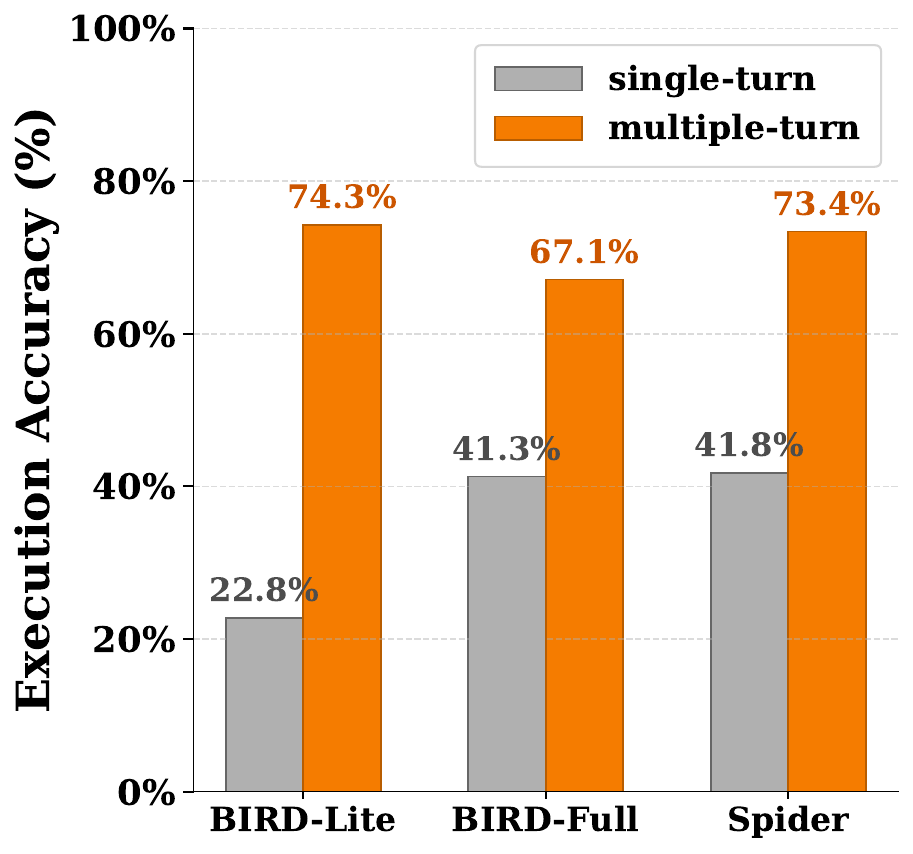}
        \caption{}
    \end{subfigure}
    \quad
    \begin{subfigure}[t]{0.35\textwidth}
        \centering
        \includegraphics[width=\linewidth]{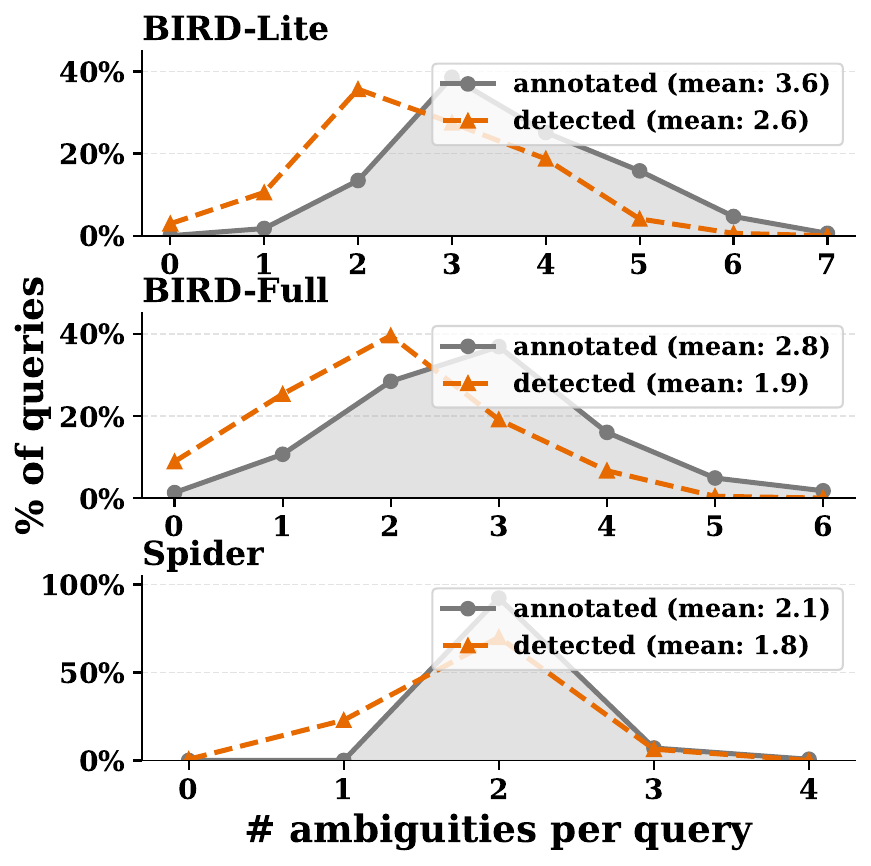}
        \caption{}
    \end{subfigure}

    \caption{
    (a) Execution accuracy comparison on three datasets in single- and multiple-turn settings.
    (b) Ground-truth annotated ambiguity distributions versus ambiguities detected through multiple-turn interaction.
    }
    \vspace{-0.5cm}
    \label{fig:k0_vs_naive}
\end{figure}

Modern LLM agents make this interaction increasingly feasible. Equipped with tools for inspecting schemas, querying database contents, and communicating with users, an agent can iteratively gather information before producing the final SQL~\cite{dinsql,chess,mint,macsql,dci}. Indeed, we find that allowing such interaction dramatically improves execution accuracy compared with single-turn generation—for example, from 22.8\% to 74.3\% on BIRD-Lite, as shown in Figure~\ref{fig:k0_vs_naive} (a). Yet high execution accuracy does not imply that the generated query is well grounded in the user’s stated intent. On the same benchmark, the agent identifies only 2.6 of 3.6 annotated ambiguities on average. Similar gaps appear on BIRD-Full and Spider, as shown in Figure~\ref{fig:k0_vs_naive} (b). In other words, an agent can sometimes produce an executable answer without explicitly resolving the underspecifications that determine that answer. Such silent assumptions are undesirable even when they happen to yield the correct execution result, as a small change in the intended interpretation could make the same behavior fail.

Why do capable LLM agents leave these ambiguities unresolved? An intuitive explanation is that they do not ask enough questions. We therefore conduct a controlled question-budget study in which the agent is instructed to ask a fixed number \(k\) of clarification questions before submitting its SQL. Increasing \(k\) substantially improves execution accuracy on BIRD-Lite, from 64.9\% at \(k=2\) to 88.9\% at \(k=10\). 
However, the additional questions exhibit sharply diminishing value: annotated ambiguities are concentrated among the earlier questions, whereas later questions increasingly concern answerable but non-critical issues that do not correspond to annotated ambiguities, or unanswerable issues for which the task specification provides no grounded answer.
Thus, simply forcing the agent to ask more questions is effective but inefficient. The central challenge is not maximizing the number of questions, but identifying and faithfully resolving the important ones.

Plan mode, where an agent first constructs a workflow before execution, has become a de facto paradigm for achieving state-of-the-art agent performance~\cite{react,hugginggpt}. A natural approach is therefore to extend this paradigm to clarification, asking the agent to plan its clarification process in advance. Surprisingly, we find that prompting alone is insufficient. When instructed to first identify and rank potential clarification questions and then execute that plan, the agent frequently abandons questions that it had already identified as relevant. On BIRD-Lite, for example, the agent plans 4.1 questions on average but asks only 3.3. Inspection of the trajectories reveals a characteristic failure pattern: after receiving a few answers or obtaining additional evidence from the database, the agent becomes prematurely confident that it has enough information and proceeds to SQL generation, silently skipping the remaining planned questions. This observation exposes an important distinction between identifying an ambiguity and maintaining a commitment to resolve it.

Motivated by this distinction, we introduce \model, an externalized clarification-planning mechanism for agentic Text-to-SQL. Instead of representing the clarification plan only as natural-language context that the model may later ignore, \model materializes it as a mutable question pool managed through explicit actions. Specifically, the agent first generates a ranked set of clarification questions after exploring the datasource. During interaction, it may ask the highest-priority question, explicitly drop a question that has become redundant or irrelevant, or add a newly discovered ambiguity. Crucially, the agent is not allowed to submit its final SQL while unresolved questions remain in the pool. Every planned clarification must therefore be explicitly discharged rather than silently forgotten. At the same time, as questions can be added or dropped as new information arrives, the mechanism remains adaptive rather than reducing clarification to a rigid checklist.

We evaluate \model on three benchmarks derived from BIRD-Interact-Lite, BIRD-Interact-Full, and Spider, using an agentic interaction framework in which the system can explore the datasource and ask a simulated user for clarification. Across all three datasets, \model consistently improves ambiguity coverage and grounded success while substantially reducing silent failures, with competitive end-to-end execution accuracy. For example, on BIRD-Lite, it improves ambiguity recall from 80.8\% with prompt-based planning to 87.5\% and reduces the silent-failure rate from 10.5\% to 8.2\%. 
The mechanism is also genuinely dynamic: on the more challenging BIRD benchmarks, roughly one third of instances invoke pool modifications after initialization. 

We summarize the main contributions of this work as follows:

\begin{itemize}[leftmargin=*]
    \item We identify premature clarification termination as a key failure mode in agentic Text-to-SQL under underspecified queries. Through controlled question-budget and planning analyses, we show that LLM agents can identify important ambiguities yet fail to follow through on resolving them, while brute-force questioning wastes interaction on increasingly low-value questions.
    \item We propose \model, an externalized and dynamically managed clarification protocol. By turning clarification intentions into explicit state that must be asked, deliberately dropped, or augmented before submission, \model prevents the silent abandonment of identified ambiguities while remaining adaptive to information revealed during interaction.
    \item We provide a systematic evaluation of correctness, groundedness, and interaction efficiency across three benchmarks. Our results show that explicit commitment management substantially improves ambiguity coverage and reduces silent failures over unconstrained interaction, post-hoc reflection, and prompt-based planning, while retaining competitive SQL execution accuracy.
\end{itemize}

\vspace{-0.2cm}
\section{Problem Formulation}
\label{sec:problem}

\vspace{-0.2cm}
\paragraph{Agentic Text-to-SQL under underspecification.}
We study Text-to-SQL in an interactive setting where the user's initial request may be underspecified.
Given a natural-language query $q$ and a datasource $D$, the goal is not necessarily to generate SQL immediately.
Instead, an agent interacts with both the datasource and the user before producing a final SQL query.
We represent an interaction trajectory as
\(
\tau = (q, a_1, o_1, \ldots, a_T, o_T),
\)
where at each step the agent may take one of three types of actions:
(1) explore the datasource through the available tools,
(2) ask the user a clarification question, or
(3) submit a SQL query, which terminates the interaction.
The agent must therefore determine not only how to implement the user's request, but also whether the information available so far is sufficient to determine the intended implementation.

\vspace{-0.2cm}
\paragraph{Underspecification and ambiguities.}
We call a query underspecified when the current interaction context admits multiple plausible interpretations that can lead to different intended SQL semantics. We refer to each unresolved piece of information as an ambiguity.
For an input query $q$, let
\(
A^\star(q) = \{z_1,\ldots,z_m\}
\)
denote the set of annotated ambiguities that require user clarification to resolve.
During an interaction $\tau$, the agent may ask clarification questions that address a subset of these ambiguities.
Let
\(
A(\tau) \subseteq A^\star(q)
\)
denote the set of annotated ambiguities addressed by the agent's clarification questions during $\tau$.
Operationally, an annotated ambiguity is considered addressed when the agent asks a corresponding clarification question on it.
The ambiguity coverage of an interaction can then be measured by
\(
\mathrm{Recall}_{\mathrm{amb}}(\tau)
=
|A(\tau)| / |A^\star(q)|.
\)

\vspace{-0.2cm}
\paragraph{Correctness versus resolution groundedness.}
Standard Text-to-SQL evaluation focuses on whether the generated SQL produces the correct execution result.
Let $\hat{s}$ be the submitted SQL and $s^\star$ the reference SQL.
We define execution accuracy as
\(
\mathrm{EA}(\hat{s},s^\star)
=
\mathbbm{1}
\left[
\operatorname{Exec}(\hat{s},D)
=
\operatorname{Exec}(s^\star,D)
\right].
\)
We distinguish execution correctness from resolution groundedness, which captures whether the agent has explicitly addressed the underspecifications underlying the intended query.
In particular, we define \emph{grounded success} as an interaction that both produces the correct execution result and addresses all annotated ambiguities:
\(
\mathrm{GS}(\tau)
=
\mathbbm{1}
\left[
(\mathrm{EA}=1)
\;\land\;
(A(\tau)=A^\star(q))
\right].
\) 
Correspondingly, a correct answer that leaves at least one ambiguity unresolved is an \emph{ungrounded success}, while an incorrect answer that leaves at least one ambiguity unresolved constitutes a \emph{silent failure}.
These distinctions allow us to separate answers that are merely correct from those whose correctness is supported by explicit resolution of the user's underspecified intent.
An effective agent should resolve the underspecifications needed for a grounded answer while avoiding unnecessary interaction and retaining strong end-to-end SQL execution accuracy.

\section{Diagnosing Premature Clarification Termination}
\label{sec:diagnosis}

Having formulated agentic Text-to-SQL under underspecification, we now examine why capable LLM agents still leave ambiguities unresolved despite having access to interactive clarification.
Our analysis reveals a recurring pattern: agents often terminate clarification before all relevant underspecification has been resolved.
We refer to this behavior as \emph{premature clarification termination}.

As previewed in Figure~\ref{fig:k0_vs_naive}, allowing an agent to interact with the user substantially improves Text-to-SQL execution accuracy, but does not eliminate unresolved underspecification.
Thus, access to clarification is not sufficient: agents often terminate the interaction while ambiguities remain unresolved.
This observation motivates the first question in our diagnosis:
is the remaining grounding gap simply a consequence of asking too few questions?

\subsection{More Questions Help, but with Diminishing Value}
\label{sec:diagnosis_budget}

\begin{figure}[!t]
    \centering

    \begin{subfigure}[t]{0.24\textwidth}
        \centering
        \includegraphics[width=\linewidth]{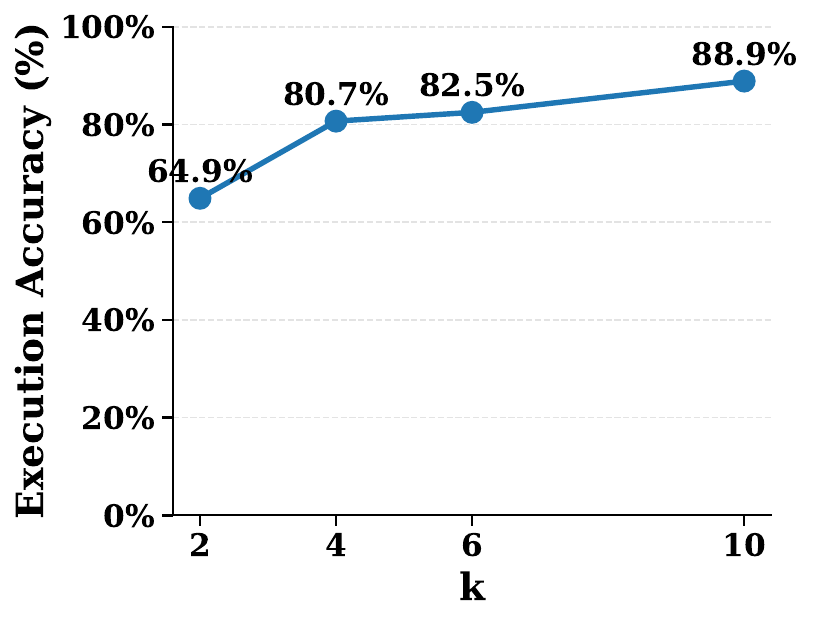}
        \caption{}
    \end{subfigure}
    \hfill
    \begin{subfigure}[t]{0.24\textwidth}
        \centering
        \includegraphics[width=\linewidth]{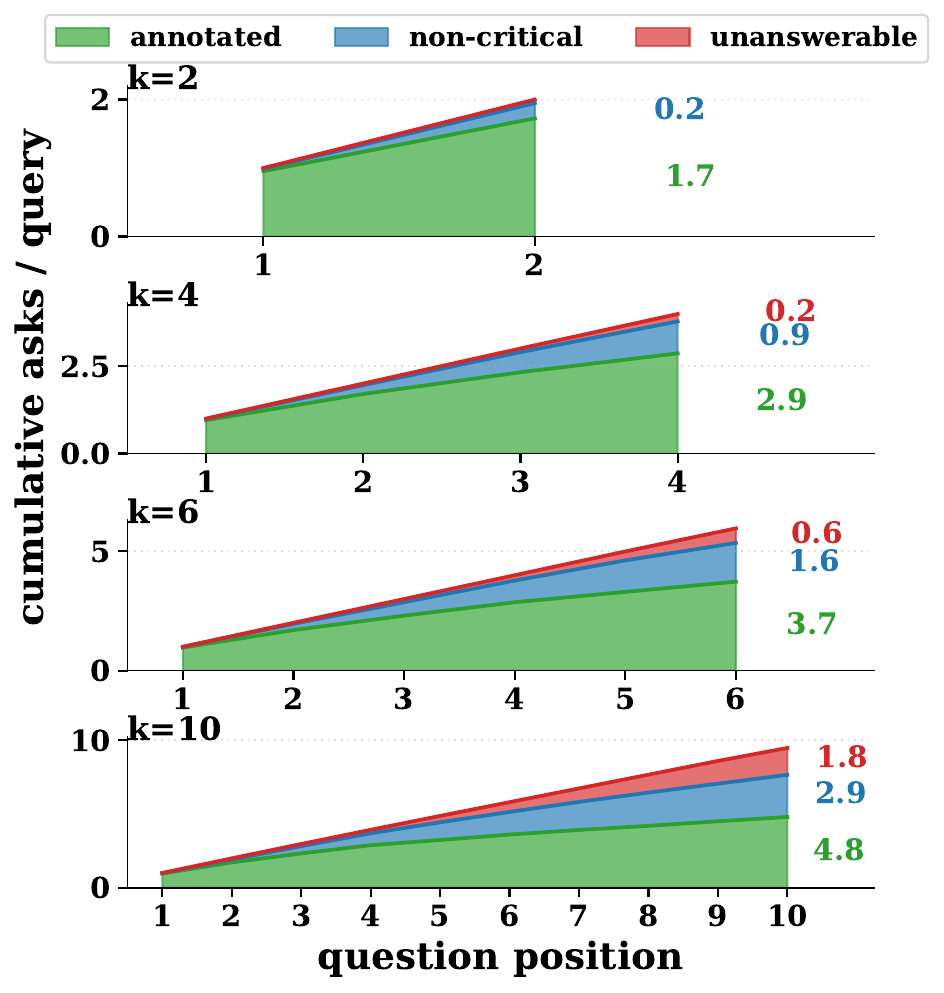}
        \caption{}
    \end{subfigure}
    \hfill
    \begin{subfigure}[t]{0.24\textwidth}
        \centering
        \includegraphics[width=\linewidth]{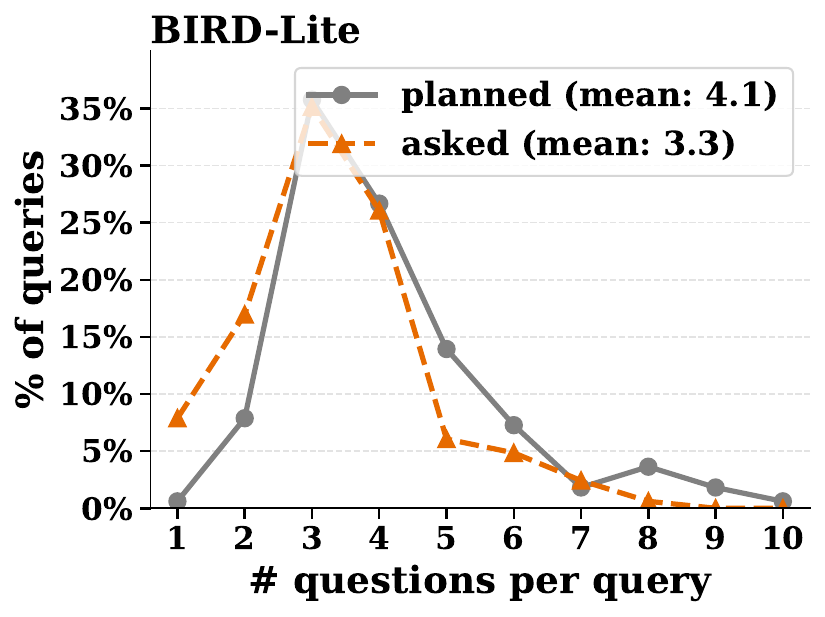}
        \caption{}
    \end{subfigure}
    \hfill
    \begin{subfigure}[t]{0.24\textwidth}
        \centering
        \includegraphics[width=\linewidth]{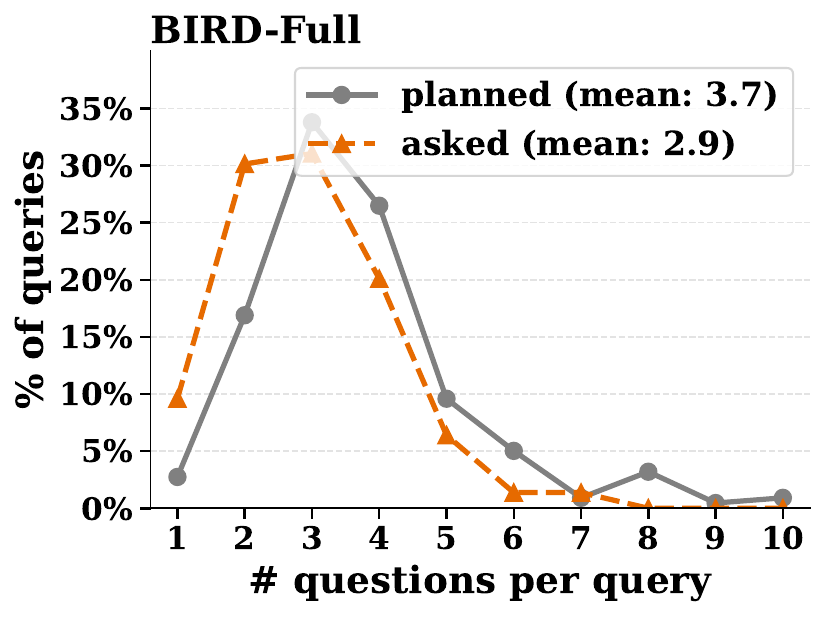}
        \caption{}
    \end{subfigure}
    \vspace{-0.3cm}
    \caption{
    (a) Execution accuracy scaling with the number of clarification questions $k$.
    (b) Cumulative composition of asked clarification questions as $k$ increases.
    (c), (d) Planned versus actually asked question distributions for prompting-based plan mode on BIRD-Lite and BIRD-Full.
    }
    \label{fig:combined_analysis}
\end{figure}

An intuitive explanation for the remaining grounding gap is that unconstrained agents simply ask too few questions.
To test this hypothesis, we conduct a controlled question-budget study in which the agent is instructed to ask $k$ clarification questions before submitting a SQL query.
We vary $k \in \{2,4,6,10\}$ on BIRD-Lite.
Due to imperfect instruction following, the realized number of questions can be slightly smaller than the target \(k\).

Figure~\ref{fig:combined_analysis}(a) shows that increasing the clarification budget indeed improves execution accuracy, where accuracy rises from 64.9\% at $k=2$ to 88.9\% at $k=10$.
This confirms that premature submission is an important source of error: forcing the agent to continue asking questions can recover substantial performance.

However, the additional interaction becomes progressively less informative.
Figure~\ref{fig:combined_analysis}(b) decomposes the questions at each budget into those matching annotated ambiguities, non-critical questions, and questions that the user cannot answer.
At $k=2$, most questions target annotated ambiguities.
As the budget grows, however, an increasing fraction of the additional questions are spent on non-critical or unanswerable issues.
At $k=10$, the agent asks 2.9 non-critical and 1.8 unanswerable questions per query on average, in addition to 4.8 questions matching annotated ambiguities. 
These results show that brute-force clarification can mitigate premature termination, but does so inefficiently.
The desired behavior is therefore not to maximize the number of questions, but to ensure that the agent follows the important ambiguities it has identified while avoiding unnecessary clarification.

\subsection{Prompted Planning Identifies Ambiguities but Fails to Follow Through}
\label{sec:diagnosis_planning}

The previous analysis suggests a natural alternative to brute-force questioning: ask the agent to identify the important ambiguities in advance and then resolve them according to a clarification plan.
We therefore consider a planning strategy by instructing the agent to first explore the datasource, generate a ranked list of clarification questions, and then ask these questions before producing the final SQL through system prompt.

However, planning improves question selection, but introduces a different failure mode.
As shown in Figure~\ref{fig:combined_analysis}(c)-(d), the number of questions actually asked is systematically smaller than the number initially planned.
In other words, explicitly identifying an ambiguity does not ensure that the agent will later resolve it.
Through inspecting the interaction trajectories, we find the agent often proceeds directly to SQL generation even though previously planned clarification questions remain after receiving several user responses or obtaining additional evidence through datasource exploration.
Those questions are therefore not explicitly judged irrelevant or resolved but simply disappear from the subsequent reasoning process.

This observation exposes a limitation of prompt-based planning.
The plan is represented only as text in the agent's conversational context, while the decision of whether to continue following it remains entirely under the model's control.
As the interaction evolves, the model may become sufficiently confident in one interpretation to stop clarification, even when its own earlier analysis identified additional unresolved issues.
Prompting can therefore help the model identify what should be clarified, but provides no mechanism that requires it to explicitly account for those questions before submission.

Through the empirical study, we find that the central challenge is not merely ambiguity detection or question generation.
Agents can identify useful clarification questions, and additional questioning can substantially improve performance.
The failure lies in reliably maintaining unresolved clarification needs throughout an evolving interaction.
This motivates an explicit mechanism in which identified questions persist as external state until they are either addressed or deliberately removed.

\section{\model: Externalizing Clarification Plans}
\label{sec:method}

The analysis in Section~\ref{sec:diagnosis} shows that the key challenge is not merely identifying clarification needs, but reliably maintaining them until they are explicitly addressed.
To this end, we introduce \model, which externalizes the agent's clarification plan as persistent, explicitly managed state.
Rather than relying on a soft instruction to follow a plan, \model turns clarification planning into an explicit interaction protocol.

\vspace{-0.3cm}
\paragraph{Pool initialization.}
Given an underspecified user query $q$ and datasource $D$, the agent first explores the datasource using the same tools as the underlying agent.
Once it considers the initial exploration sufficient, it generates a ranked clarification plan
\(
P_0 = [z_1, z_2, \ldots, z_m],
\)
where each $z_i$ is a clarification question and the ordering reflects the agent's estimated priority.
Unlike prompt-based planning, where the plan remains ordinary text in the conversational context, \model maintains the current pool $P_t$ as explicit external state throughout the subsequent interaction.
As a result, a previously identified clarification need cannot silently disappear simply because it is no longer salient in the model's current reasoning.

\vspace{-0.3cm}
\paragraph{Dynamic pool management.}
The pool is updated as the interaction evolves.
New user responses or datasource observations may make previously planned questions no longer necessary to ask, for example because the relevant information has already been addressed or the question has become redundant or irrelevant. They may also reveal ambiguities that were not apparent during initialization.
Accordingly, the agent may inspect the current pool, remove questions that no longer require clarification, add newly discovered questions, and select pending questions to ask.
This makes the clarification plan adaptive rather than a static checklist, while keeping unresolved clarification needs explicitly represented throughout the interaction.

\vspace{-0.3cm}
\paragraph{Explicit accounting before submission.}
The agent may submit the final SQL only when the clarification pool is empty,
\(
P_t = \varnothing.
\)
Importantly, this does not require every initially planned question to be asked separately: a question may be explicitly removed when prior interaction has already addressed the information it seeks, or when subsequent evidence makes it redundant or irrelevant.
Conversely, newly discovered clarification needs may be added to the pool and become subject to the same requirement.
Thus, \model enforces explicit accounting rather than exhaustive questioning: the agent remains free to revise its plan, but every pending item must be addressed or deliberately removed before submission.
Once the pool is empty, the agent generates the final SQL using the full interaction context.
The underlying language model, datasource tools, and SQL-generation procedure are otherwise unchanged.

\section{Experiments}
\label{sec:experiments}

\subsection{Experimental Setup}
\label{sec:experimental_setup}

\paragraph{Interaction framework.}
We evaluate all methods in the same agentic Text-to-SQL environment, consisting of a system agent and a user simulator.
Given an underspecified query and access to the corresponding datasource, the system agent may iteratively explore the datasource, ask the user clarification questions, and eventually submit a SQL query.
For the user simulator, we directly adopt the function-driven user-simulation protocol from~\cite{birdinteract}, which was designed to mitigate ground-truth leakage and has been independently validated for reliability and alignment with human-user interactions. Specifically, the simulator first determines whether a clarification question can be answered from the task specification and, when answerable, generates a natural-language response conditioned on the corresponding task-grounded reference information.
We use Claude-Opus-4.8~\cite{opus48} as the backbone LLM for both the system agent and the user simulator.
Across all compared methods, we keep the underlying language model, datasource tools, user simulator, and SQL-generation environment fixed; the methods differ only in how they manage clarification.

\begin{wraptable}[6]{r}{0.55\textwidth}
  \vspace{-25pt}
  \centering
  \caption{Dataset statistics.}\label{tab:dataset-stats}
  \vspace{-5pt}
  \scalebox{0.8}{
\begin{tabular}{lccc}
\toprule
\textbf{Statistic} & \textbf{BIRD-Lite} & \textbf{BIRD-Full} & \textbf{Spider} \\
\midrule
\# Instances              & 171   & 225   & 297  \\
\# Distinct Databases     & 18    & 22    & 19   \\
\# Tokens / User Query    & 31.8  & 24.0  & 14.3 \\
\# Tokens / SQL           & 149.5 & 78.5  & 18.1 \\
\# Ambiguities / Task     & 3.56  & 2.77  & 2.08 \\
\bottomrule
\end{tabular}
}
\end{wraptable}
\paragraph{Dataset construction.}
We evaluate on three benchmarks with multiple annotated ambiguities per query: two subsets derived from BIRD-Interact and one benchmark constructed from Spider.
Their statistics are summarized in Table~\ref{tab:dataset-stats}.

\begin{itemize}[leftmargin=*]
\item \textbf{BIRD-Interact.}
BIRD-Interact~\cite{birdinteract} provides underspecified Text-to-SQL queries with annotated intent, implementation, and knowledge-level ambiguities.
Following prior observations~\cite{twosurface} that many implementation-level variations are normalized away by execution-based evaluation on this benchmark, we only focus on intent and knowledge-level ambiguities that can critically influence the results.
To isolate ambiguity handling from failures caused purely by intrinsic SQL-generation difficulty, we further retain only instances that the system agent can solve when all annotated clarifications are provided in advance.
Specifically, we generate natural-language responses for every annotated ambiguity, inject the resulting clarification exchanges into the interaction history, and keep an instance only if the agent subsequently produces the correct execution result.
This yields 171 of 300 instances from BIRD-Interact-Lite and 225 of 600 from BIRD-Interact-Full, which we refer to as BIRD-Lite and BIRD-Full, respectively.
Thus, our evaluation intentionally measures ambiguity handling conditional on the task being solvable once all annotated ambiguities are resolved. This solvability filter prevents intrinsic SQL-generation failures from confounding comparisons among clarification strategies. More details are in Appendix~\ref{app:bird_filtering}.

\item \textbf{Spider.}
Existing ambiguous Text-to-SQL benchmarks typically contain only a small number of ambiguities per query.
To obtain a complementary benchmark with multiple controlled ambiguities, we extend the corruption procedure used by PRACTIQ~\cite{practiq} to introduce multiple underspecifications into Spider queries.
The construction considers ambiguity in selected columns, filtering columns, cell values, and filtering criteria, while updating the datasource and reference SQL accordingly and validating the resulting instance by execution.
We retain 297 valid instances and refer to this benchmark as Spider~\cite{spider}.
The full construction pipeline is provided in Appendix~\ref{app:spider_construction}.
\end{itemize}

\paragraph{Compared methods.}
We compare \model against six baselines representing three canonical strategies for inference-time clarification.
Rather than reproducing the original task-specific systems, we instantiate the core ideas of these strategies within the same agentic Text-to-SQL harness, keeping the underlying model, tools, and user simulator fixed.
(1) \emph{Unstructured methods}, including \textbf{Naive}, a ReAct-style agent~\cite{react} that freely decides when to explore, clarify, or submit; \textbf{Read-All}, a full-context control that exposes all datasource information upfront; \textbf{Regenerate}, motivated by consistency-based inference~\cite{selfconsistency,manakul2023selfcheckgpt}, which generates an independent second SQL candidate and resumes clarification when the two disagree; and \textbf{Draft}, inspired by draft-then-verify approaches~\cite{dhuliawala2024chain}, which first constructs a best-effort SQL before clarification.
(2) \emph{Post-hoc reasoning method}, \textbf{Self-Reflection}, following the general self-refinement paradigm~\cite{madaan2023self,shinn2023reflexion}, audits the proposed SQL for unverified assumptions before submission and continues clarification when necessary.
And (3) \emph{explicit planning methods}, including \textbf{Plan-Prompt}, motivated by plan-then-execute prompting~\cite{wang2023plan}, which generates and ranks a clarification plan before execution, and our \textbf{\model}, which externalizes that plan as persistent clarification state.

\paragraph{Evaluation metrics.}
We evaluate both end-to-end correctness and clarification groundedness.
Following Section~\ref{sec:problem}, we report \emph{Execution Accuracy} (EA), \emph{Ambiguity Recall}, \emph{Grounded Success} (GS), \emph{Ungrounded Success} (US), \emph{Silent Failure} (SF), and the average number of clarification questions (\#Asks) as a measure of interaction cost.
For EA, we use the original execution evaluators provided by BIRD-interact and Spider for their respective instances.
For cost analysis, we also record input/output token usage, and monetary cost based on AWS Bedrock. 
\begin{table}[t]
\centering
\caption{Method comparison across datasets.
}
\label{tab:main_results}
\scalebox{0.75}{
\begin{tabular}{llcccccc}
\toprule
\textbf{Dataset} & Method & EA $\uparrow$ & Recall $\uparrow$ & GS $\uparrow$
& US $\downarrow$ & SF $\downarrow$ & \#Asks $\downarrow$ \\
\midrule

\multirow{7}{*}{BIRD-Lite}
  & Naive
  & 74.3 & 73.9 & 33.9 & 40.4 & 17.0 & 2.3 \\
  & Read-All
  & 69.6 & 70.8 & 29.8 & 39.8 & 18.7 & \textbf{2.0} \\
  & Regenerate
  & 74.9 & 73.7 & 31.6 & 43.3 & 15.2 & 2.3 \\
  & Draft
  & 75.4 & 71.6 & 28.7 & 46.8 & 19.9 & 2.5 \\
  & Self-Reflection
  & 77.2 & 75.5 & 35.7 & 41.5 & 16.4 & 2.7 \\
  & Plan-Prompt
  & 79.5 & 80.8 & 42.1 & 37.4 & 10.5 & 3.3 \\
  & \cellcolor{light-gray}\textbf{\model}
  & \cellcolor{light-gray}\textbf{81.3}
  & \cellcolor{light-gray}\textbf{87.5}
  & \cellcolor{light-gray}\textbf{49.1}
  & \cellcolor{light-gray}\textbf{32.2}
  & \cellcolor{light-gray}\textbf{8.2}
  & \cellcolor{light-gray}4.2 \\

\midrule

\multirow{7}{*}{BIRD-Full}
  & Naive
  & 67.1 & 68.6 & 31.6 & 35.6 & 19.1 & 2.0 \\
  & Read-All
  & 62.7 & 70.2 & 33.3 & 29.3 & 17.3 & \textbf{1.7} \\
  & Regenerate
  & 61.3 & 69.9 & 35.1 & 26.2 & 15.6 & 2.0 \\
  & Draft
  & 65.3 & 68.4 & 33.3 & 32.0 & 18.7 & 2.2 \\
  & Self-Reflection
  & \textbf{70.7} & 71.3 & 33.8 & 36.9 & 18.7 & 2.3 \\
  & Plan-Prompt
  & 69.8 & 78.8 & 45.8 & 24.0 & 11.1 & 2.9 \\
  & \cellcolor{light-gray}\textbf{\model}
  & \cellcolor{light-gray}67.1
  & \cellcolor{light-gray}\textbf{84.6}
  & \cellcolor{light-gray}\textbf{48.0}
  & \cellcolor{light-gray}\textbf{19.1}
  & \cellcolor{light-gray}\textbf{8.4}
  & \cellcolor{light-gray}3.6 \\

\midrule

\multirow{7}{*}{Spider}
  & Naive
  & 73.4 & 87.4 & 58.9 & 14.5 & 7.1 & 1.9 \\
  & Read-All
  & 68.7 & 81.6 & 49.5 & 19.2 & 12.8 & \textbf{1.6} \\
  & Regenerate
  & 72.1 & 87.9 & 56.6 & 15.5 & 5.7 & 1.9 \\
  & Draft
  & 74.1 & 88.2 & 59.9 & 14.1 & 6.7 & 2.0 \\
  & Self-Reflection
  & 75.1 & 91.8 & 66.3 & 8.8 & 5.7 & 2.2 \\
  & Plan-Prompt
  & 75.4 & 93.2 & 69.0 & 6.4 & 4.7 & 2.2 \\
  & \cellcolor{light-gray}\textbf{\model}
  & \cellcolor{light-gray}\textbf{76.1}
  & \cellcolor{light-gray}\textbf{94.3}
  & \cellcolor{light-gray}\textbf{70.7}
  & \cellcolor{light-gray}\textbf{5.4}
  & \cellcolor{light-gray}\textbf{4.0}
  & \cellcolor{light-gray}2.5 \\

\bottomrule
\end{tabular}
}
\end{table}

\subsection{Main Results: Correctness and Groundedness}
\label{sec:main_results}

Table~\ref{tab:main_results} reports the main results across the three benchmarks.
Overall, \model achieves the strongest groundedness performance on all datasets.
Compared with Plan-Prompt, its closest planning-based baseline, \model consistently improves ambiguity recall and grounded success while reducing both ungrounded success and silent failure.
On BIRD-Lite, for example, ambiguity recall increases from 80.8\% to 87.5\%, grounded success from 42.1\% to 49.1\%, and silent failure decreases from 10.5\% to 8.2\%.
The same pattern holds on BIRD-Full and Spider, suggesting that externalizing the clarification plan is more effective than relying on the model to follow a plan represented only in conversational context.

We also notice that Read-All provides a complementary diagnostic.
Despite receiving all datasource information upfront, it consistently asks the fewest clarification questions and underperforms the planning-based methods in both correctness and groundedness.
This behavior is consistent with premature clarification termination: additional evidence alone does not resolve underspecification and may instead lead the agent to submit once it becomes confident that it has sufficient information.

Importantly, stronger groundedness does not necessarily translate into higher execution accuracy.
On BIRD-Full, Self-Reflection achieves the highest EA of 70.7\%, compared with 67.1\% for \model, yet its ambiguity recall and grounded success are substantially lower.
The decomposition into grounded and ungrounded successes makes this distinction particularly clear: 36.9\% of instances under Self-Reflection are correct despite leaving at least one annotated ambiguity unresolved, compared with only 19.1\% for \model.
This result illustrates why execution accuracy alone is insufficient for underspecified queries.
Post-hoc reconsideration can recover correct SQL in some cases, but it does not substitute for persistently maintaining and resolving clarification needs throughout the interaction.

\subsection{How Does \model Change Clarification Behavior?}
\label{sec:behavior_analysis}

We next examine how \model changes the agent's clarification behavior beyond the aggregate performance reported above.

\begin{wraptable}[9]{r}{0.55\textwidth}
  \vspace{-10pt}
  \centering
\caption{\model tool usage across datasets.}\label{tab:pool-usage}
\scalebox{0.8}{
\begin{tabular}{lccc}
\toprule
Statistic & BIRD-Lite & BIRD-Full & Spider \\
\midrule
Initial Pool Size          & 3.9 & 3.3 & 2.3 \\
Actual Asks           & 4.2 & 3.6 & 2.5 \\
Unmodified Pool                   & 65.5\% & 61.8\% & 81.5\% \\
Adding question usage     & 30.4\% & 32.9\% & 18.5\% \\
Dropping question usage    & 4.7\% & 10.7\% & 0.7\% \\
\bottomrule
\end{tabular}
}
\end{wraptable}
\paragraph{The clarification plan is actively revised.}
Table~\ref{tab:pool-usage} summarizes how the question pool evolves during interaction.
Although the majority of initial pools can be executed without modification, the agent frequently revises the plan on more challenging instances.
On BIRD-Lite and BIRD-Full, 34.5\% and 38.2\% of pools are modified, respectively, compared with 18.5\% on Spider.
Moreover, adding question is used substantially more often than dropping question: the agent adds newly discovered clarification needs in 30.4\% and 32.9\% of BIRD-Lite and BIRD-Full instances, while dropping existing questions in only 4.7\% and 10.7\%.
Consequently, the number of questions actually asked is slightly larger than the initial pool size on all three datasets.
These results show that \model does not simply enforce a static initial checklist; instead, the clarification plan is actively adapted as new information becomes available.

\begin{figure}[h]
    \centering

    \begin{subfigure}[t]{0.32\textwidth}
        \centering
        \includegraphics[width=\linewidth]{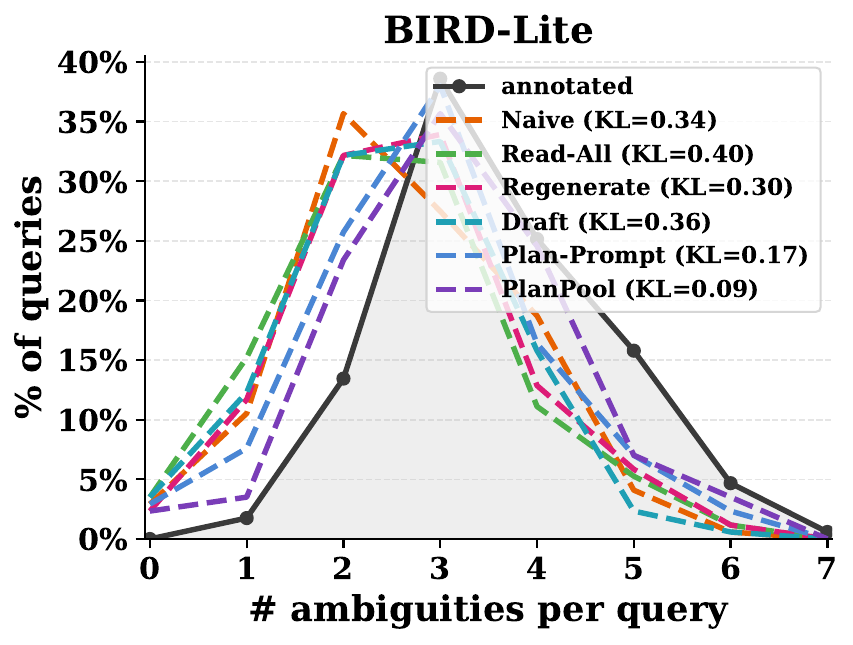}
        \caption{}
    \end{subfigure}
    \hfill
    \begin{subfigure}[t]{0.32\textwidth}
        \centering
        \includegraphics[width=\linewidth]{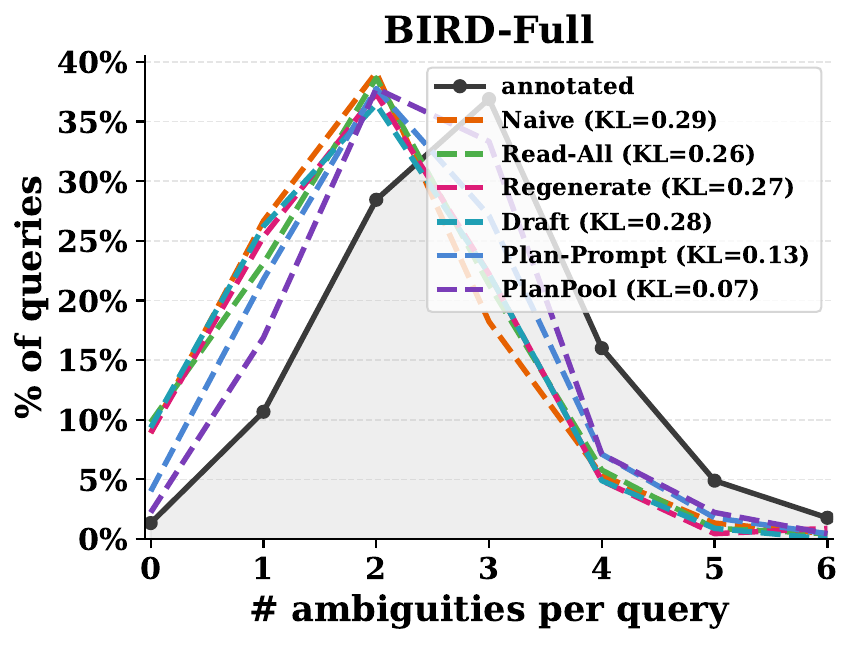}
        \caption{}
    \end{subfigure}
    \hfill
    \begin{subfigure}[t]{0.32\textwidth}
        \centering
        \includegraphics[width=\linewidth]{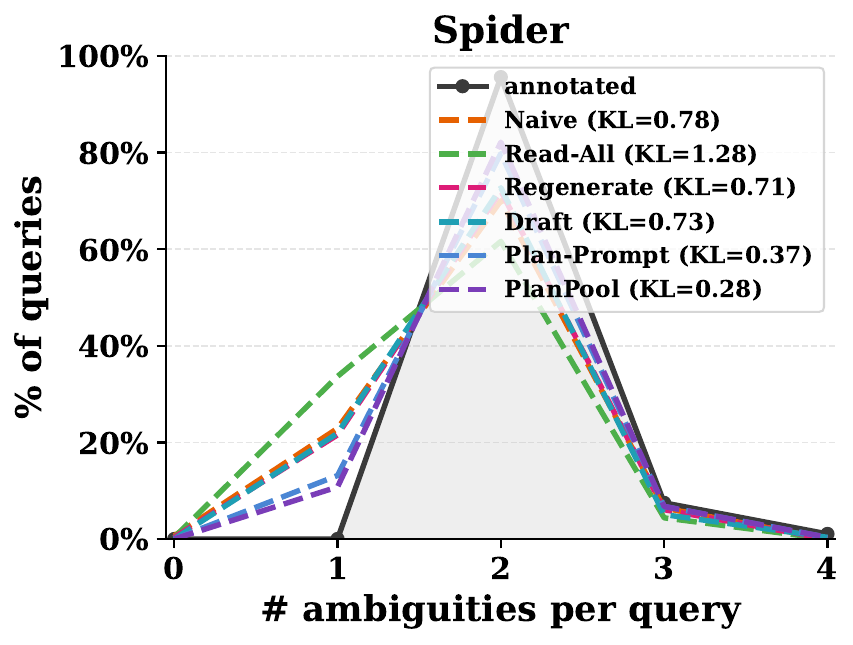}
        \caption{}
    \end{subfigure}

    \caption{
    Annotated versus detected ambiguity distributions produced by different methods, together with the KL divergence with respect to the annotated ground-truth distribution:
    (a) BIRD-Lite, (b) BIRD-Full, and (c) Spider. 
    }
    \label{fig:identified_vs_annotated}
\end{figure}

\vspace{-0.5cm}
\paragraph{The mismatch in ambiguity coverage is substantially reduced.}
Figure~\ref{fig:identified_vs_annotated} compares the distribution of detected ambiguity counts with the annotated distribution.
Prompt-based planning already narrows this gap relative to the naive agent, but \model consistently yields the closest match across all three datasets.
Specifically, the KL divergence decreases from 0.34 for Naive and 0.17 for Plan-Prompt to 0.09 for \model on BIRD-Lite; from 0.29 and 0.13 to 0.07 on BIRD-Full; and from 0.78 and 0.37 to 0.28 on Spider.
Together with the ambiguity-recall improvements in Table~\ref{tab:main_results}, this suggests that \model not only recovers more of the annotated ambiguities, but also substantially reduces the systematic under-clarification.

\begin{table}[t]
\centering
\caption{Computational cost across datasets. Token counts are averaged per query.}
\label{tab:cost}
\scalebox{0.8}{
\begin{tabular}{llccc}
\toprule
\textbf{Dataset} & Method & Input Tok. $\downarrow$ & Output Tok. $\downarrow$ & Cost (\$) $\downarrow$ \\
\midrule

\multirow{6}{*}{BIRD-Lite}
  & Naive
  & 136k & 2.4k & 0.36 \\
  & Read-All
  & 122k & 1.8k & 0.30 \\
  & Regenerate
  & 169k & 3.1k & 0.35 \\
  & Draft
  & 139k & 2.7k & 0.29 \\
  & Plan-Prompt
  & 148k & 2.6k & 0.29 \\
  & \cellcolor{light-gray}\textbf{\model}
  & \cellcolor{light-gray}201k
  & \cellcolor{light-gray}2.5k
  & \cellcolor{light-gray}0.32 \\

\midrule

\multirow{6}{*}{BIRD-Full}
  & Naive
  & 139k & 2.6k & 0.27 \\
  & Read-All
  & 132k & 2.0k & 0.30 \\
  & Regenerate
  & 168k & 3.3k & 0.37 \\
  & Draft
  & 146k & 2.9k & 0.29 \\
  & Plan-Prompt
  & 150k & 2.7k & 0.29 \\
  & \cellcolor{light-gray}\textbf{\model}
  & \cellcolor{light-gray}184k
  & \cellcolor{light-gray}2.6k
  & \cellcolor{light-gray}0.30 \\

\midrule

\multirow{6}{*}{Spider}
  & Naive
  & 20k & 1.0k & 0.07 \\
  & Read-All
  & 13k & 0.7k & 0.05 \\
  & Regenerate
  & 25k & 1.3k & 0.08 \\
  & Draft
  & 23k & 1.2k & 0.07 \\
  & Plan-Prompt
  & 23k & 1.2k & 0.07 \\
  & \cellcolor{light-gray}\textbf{\model}
  & \cellcolor{light-gray}28k
  & \cellcolor{light-gray}1.1k
  & \cellcolor{light-gray}0.07 \\

\bottomrule
\end{tabular}
}
\end{table}

\vspace{-0.2cm}
\subsection{Interaction and Computational Cost}
\label{sec:cost}
\vspace{-0.2cm}

The improved groundedness of \model comes with additional interaction.
As shown in Table~\ref{tab:main_results}, \model asks more clarification questions than the unconstrained and prompt-based baselines.
This increase is expected: \model explicitly prevents unresolved clarification needs from being silently skipped.
However, the interaction remains substantially below the larger brute-force question budgets studied in Section~\ref{sec:diagnosis_budget}, where later questions increasingly become non-critical or unanswerable.

We further report token usage and monetary cost in Table~\ref{tab:cost}.
Methods with longer interactions naturally consume more input tokens because each agentic turn re-encodes the accumulated conversation history.
Interestingly, Read-All uses the fewest tokens despite loading the datasource information upfront.
This is partly because the databases in our benchmarks are relatively compact, making the one-time read cost modest, but more importantly because Read-All tends to terminate the interaction earlier and ask fewer clarification questions.
Its low computational cost should therefore not be interpreted as greater efficiency: the shorter interaction coincides with weaker execution accuracy and groundedness, consistent with the premature-termination behavior discussed above.
In monetary terms, the differences across methods are considerably smaller than the raw token counts suggest, as repeated context is served through prompt caching at a reduced rate.
Overall, \model trades a moderate increase in interaction and token usage for substantially stronger ambiguity coverage and fewer silent failures.

\section{Related Work}
\vspace{-0.2cm}

\paragraph{Text-to-SQL.}
Text-to-SQL has evolved from supervised semantic parsing on benchmarks such as Spider~\cite{spider} to LLM-based methods that leverage decomposition, self-correction, and agent collaboration~\cite{dinsql,macsql}, with recent benchmarks further emphasizing large databases and realistic enterprise workflows~\cite{bird,spider2}.
These methods generally assume a sufficiently specified user request and focus on downstream SQL generation rather than interactive ambiguity resolution.
Recent work has studied Text-to-SQL through multi-turn interaction~\cite{sparc,cosql}, as well as datasets and methods explicitly targeting ambiguous or underspecified requests~\cite{ambigqa,ambiqt,ambrosia,chen2025learning,practiq,ambisql}, while BIRD-Interact evaluates dynamic interactions in which agents may solicit clarification and explore the database environment~\cite{birdinteract}.

\vspace{-0.2cm}
\paragraph{LLM agents.}
LLM agents interleave language-model reasoning with actions and external observations~\cite{react}, with planning playing a central role in long-horizon execution~\cite{wang2023plan,rewoo}. Other work improves agent behavior through post-hoc feedback and self-refinement~\cite{madaan2023self,shinn2023reflexion,critic}, including its application to the Text-to-SQL domain~\cite{mint}.
Recent work has increasingly emphasized that reliable agent behavior depends not only on generating plans, but also on maintaining structured information across extended trajectories using agent memory~\cite{awm,amem,cfgm,compass}.
\model is complementary to this line of research, where identified ambiguities remain explicitly represented until they are addressed or deliberately removed to prevent them from being silently abandoned before submission.

\vspace{-0.2cm}
\section{Conclusion}
\vspace{-0.2cm}

We study agentic Text-to-SQL under underspecified user queries and identify a failure mode that is not captured by execution accuracy alone: an agent may produce the correct SQL while leaving the assumptions underlying that answer unresolved.
Our analysis further shows that the central challenge is not only identifying useful clarification questions, but maintaining a commitment to unresolved information throughout an evolving interaction.
Motivated by this distinction, we introduce \textsc{PlanPool}, which externalizes clarification needs as persistent, dynamically managed state rather than leaving them solely in the model's conversational context.
Across three benchmarks, this explicit state management consistently improves ambiguity coverage and grounded success while reducing silent failures, with competitive execution accuracy.
More broadly, our results suggest that reliable agentic reasoning requires not only generating good plans, but also explicitly maintaining and discharging the unresolved commitments induced by those plans.

\subsection*{AI use statement}

In this work, we used generative AI tools both as components of the experimental framework and as assistants in the research and writing workflow.
In the experiments, LLMs were used as the system agent and the user simulator in our agentic Text-to-SQL framework, and were also used to generate natural-language clarification responses during benchmark construction and filtering.
We additionally used generative AI tools to provide feedback on experimental design and interpreting experimental results.

For manuscript preparation, we used generative AI tools to organize the paper structure, polish portions of the manuscript for readability, and assist with formatting presentation artifacts of tables, figures, and algorithms.
We did not use generative AI to formulate or prove mathematical claims, as proof-related tasks are not applicable to this work.

All AI-assisted outputs were reviewed by the authors. Experimental procedures and prompts were manually inspected. For benchmark construction and filtering steps involving LLMs, the resulting instances were further checked using the execution-based validation procedures described in the paper. AI-assisted manuscript text was reviewed and revised by the authors, and relevant citations and factual claims were checked against their original sources. We take responsibility for the final content of this work, including text, claims, code, data, and artifacts produced with the aid of generative AI.




\subsection*{Reproducibility statement}

We provide the implementation details needed to reproduce our experiments throughout the paper and appendix.
Specifically, Section~\ref{sec:experiments} describes the experimental setup, datasets, compared methods, and evaluation metrics.
Appendix~\ref{app:benchmark} details the construction and filtering procedures for the BIRD- and Spider-based benchmarks.
Appendix~\ref{app:harness} specifies the shared agentic interaction framework, datasource interaction tools, exact method-specific prompts, and user-simulation procedure.
Appendix~\ref{app:method_details} provides the execution protocols for \model and the harness-level baselines.
Together, these materials specify the data processing, interaction environment, prompting conditions, and evaluation procedures used in our experiments.

\bibliography{iclr2027_conference}
\bibliographystyle{iclr2027_conference}

\clearpage
\appendix

\section{Benchmark Construction and Processing}
\label{app:benchmark}

\subsection{BIRD-Interact Filtering}
\label{app:bird_filtering}

BIRD-Interact~\cite{birdinteract} is an ambiguous Text-to-SQL benchmark with two datasets, BIRD-Interact-Lite and BIRD-Interact-Full, containing 300 and 600 instances, respectively. Each query is associated with multiple annotated ambiguities, including intent-level ambiguities arising from vague user intent, implementation-level ambiguities concerning underspecified SQL implementation choices, and knowledge-level ambiguities caused by missing domain information. Each instance also includes a priority question and a follow-up question. For each instance, we retain only the priority question, as our evaluation focuses on the clarification most directly relevant to determining the intended SQL query. As many implementation-level variations are normalized away by execution-based evaluation~\cite{twosurface}, we only focus on intent-level and knowledge-level ambiguities that materially affect the intended query.

To isolate ambiguity handling from failures caused purely by SQL generation, we further filter the benchmarks using an oracle clarification procedure. For each instance, we first generate a natural-language clarification for every annotated critical ambiguity using the user simulator. We then construct a synthetic interaction history in which the agent has already asked about each ambiguity and received the corresponding clarification:
\begin{quote}
\small
\ttfamily
\textbf{Agent:} What do you mean by [ambiguity\textsubscript{1}]? \\
\textbf{User:} It means [clarification\textsubscript{1}]. \\
\textbf{Agent:} What do you mean by [ambiguity\textsubscript{2}]? \\
\textbf{User:} It means [clarification\textsubscript{2}]. \\
\textnormal{$\ldots$}
\end{quote}
This conversation is provided to the system agent as pre-filled interaction history, after which the agent continues normally and generates the final SQL. We retain an instance only if the resulting SQL passes execution-based evaluation. 
This filtering ensures that the retained instances are solvable by the system agent once all annotated critical ambiguities have been resolved. It therefore allows our experiments to focus on whether a clarification strategy successfully elicits the missing information, rather than on SQL-generation failures that persist even under oracle clarification.
The procedure retains 171 of 300 instances from BIRD-Interact-Lite and 225 of 600 instances from BIRD-Interact-Full. We refer to the resulting benchmarks as BIRD-Lite and BIRD-Full, respectively.

\subsection{Spider Multi-Ambiguity Construction}
\label{app:spider_construction}

We extend the ambiguity-corruption procedure of PRACTIQ~\cite{practiq}, which constructs ambiguous Text-to-SQL instances from originally unambiguous Spider examples. PRACTIQ introduces a single ambiguity into each instance; in contrast, our goal is to construct examples containing multiple controlled underspecifications. We therefore reuse its ambiguity corruption operators and apply multiple applicable operators to the same Spider instance.

Starting from a Spider instance consisting of a natural-language question, reference SQL, and database, we consider the following four corruption types:

\begin{itemize}[leftmargin=*]
    \item \textbf{Ambiguous SELECT column.}
    A selected column is replaced by two semantically similar candidate columns, making the intended \texttt{SELECT} target ambiguous.

    \item \textbf{Ambiguous values within a column.}
    A textual value appearing in a filtering condition is replaced by two semantically similar alternatives from the same column, making the intended cell value ambiguous.

    \item \textbf{Ambiguous WHERE column.}
    A filtering column is replaced by two semantically similar candidate columns that both contain the relevant value, making the intended \texttt{WHERE} column ambiguous.

    \item \textbf{Ambiguous filter criterion.}
    The natural-language description of an eligible filtering condition is rewritten so that the criterion itself becomes underspecified.
\end{itemize}

For the implementation of each individual corruption operator, including candidate selection and query rewriting, we follow PRACTIQ~\cite{practiq}. Our extension lies in composing these operators within a single instance. For each Spider example, we apply all applicable corruption operators and merge their modifications to the natural-language query, schema or cell values, and reference SQL.

\begin{figure*}[t]
\centering
\resizebox{0.96\textwidth}{!}{%
\begin{tikzpicture}[
    node distance=0.35cm,
    box/.style={
        rectangle, rounded corners=2pt, draw=black,
        fill=white, text width=3.0cm,
        minimum height=1.8cm, align=center, font=\small
    },
    io/.style={
        rectangle, draw=black, fill=white,
        text width=2.6cm, minimum height=1.0cm,
        align=center, font=\small\bfseries
    },
    arr/.style={-{Stealth[length=2.5mm]}, thick}
]

\node[io] (input) {Input\\Spider instance};

\node[box, right=of input] (s1) {%
\textbf{Ambiguous\\SELECT Column}\\[2pt]
Replace one selected column with two similar candidates.};

\node[box, right=of s1] (s2) {%
\textbf{Ambiguous\\WHERE Column}\\[2pt]
Replace one filtering column with two candidate columns.};

\node[box, right=of s2] (s3) {%
\textbf{Ambiguous Values\\Within Column}\\[2pt]
Replace one filter value with two similar candidate values.};

\node[box, right=of s3] (s4) {%
\textbf{Ambiguous Filter\\Criterion}\\[2pt]
Rewrite the question to make the criterion underspecified.};

\node[box, right=of s4] (val) {%
\textbf{Validation \&\\Annotation}\\[2pt]
Update SQL/database, execute, and extract ambiguity annotations.};

\node[io, right=of val] (output) {Output\\Spider benchmark};

\draw[arr] (input) -- (s1);
\draw[arr] (s1) -- (s2);
\draw[arr] (s2) -- (s3);
\draw[arr] (s3) -- (s4);
\draw[arr] (s4) -- (val);
\draw[arr] (val) -- (output);

\end{tikzpicture}%
}
\caption{Construction pipeline for the multi-ambiguity Spider benchmark. Each applicable operator introduces one type of underspecification, after which the modified instance is validated and annotated.}
\label{fig:spiderpipeline}
\end{figure*}
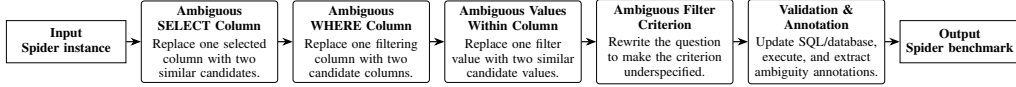

After composition, we execute the modified reference SQL on the corrupted database and discard instances that fail execution-based validation.
Since each ambiguity is introduced by a known corruption operator, its target and type are determined directly by the corresponding transformation. We use an LLM only to locate the associated textual span and SQL snippet and convert each known injected ambiguity into a unified annotation format.
The final benchmark record contains the corrupted database, ambiguous user query, validated reference SQL, and the resulting set of annotated ambiguities. This procedure yields 297 valid instances, which we refer to as Spider throughout the paper. Figure~\ref{fig:spiderpipeline} summarizes the complete construction pipeline.

\section{Agentic Interaction Framework}
\label{app:harness}

Here we describe implementation details of the harness and user simulator that are not covered in Section~\ref{sec:experimental_setup}, which are implemented following~\cite{birdinteract}.

\paragraph{Harness of the system agent.}
We directly adopt a shared ReAct~\cite{react}-style base prompt from BIRD-Interact for all methods. The prompt explicitly encourages datasource exploration and user clarification, while leaving the interaction trajectory to the agent.
The system agent controls how to inspect the datasource, ask the user for clarification, or submit its SQL, rather than following a hand-crafted workflow.
Method-specific strategies are implemented as modifications on top of this shared base harness. For \model, the base \texttt{ask(question)} action is masked at the harness level, and user clarification is instead mediated exclusively through \texttt{pool\_next()}.
The complete base system prompt used in our experiments is shown in Figure~\ref{fig:base_prompt_1} and Figure~\ref{fig:base_prompt_2}, and the method-specific system prompts in Figure~\ref{fig:prompt_readall} to Figure~\ref{fig:prompt_planpool}.

\begin{table}[h]
\centering
\caption{Datasource interaction tools adopted from BIRD-Interact. We use the original tool interfaces but ignore the action-level costs defined in BIRD-Interact.}
\label{tab:datasource_tools}
\small
\setlength{\tabcolsep}{5pt}
\begin{tabular}{p{0.48\linewidth} p{0.45\linewidth}}
\toprule
\textbf{Action} & \textbf{Observation} \\
\midrule
\texttt{execute} &
SQL execution result \\

\texttt{get\_schema} &
Database schema with example data \\

\texttt{get\_all\_column\_meanings} &
Meanings of all columns \\

\texttt{get\_column\_meaning} &
Meaning of the specified column \\

\texttt{get\_all\_external\_knowledge\_names} &
Names of available external knowledge \\

\texttt{get\_knowledge\_definition} &
Definition of the specified knowledge \\

\texttt{get\_all\_knowledge\_definitions} &
All external knowledge definitions \\

\texttt{probe\_value} &
Exact-match status and similar values \\
\bottomrule
\end{tabular}
\end{table}

\paragraph{Datasource interaction tools.}
We directly adopt the datasource interaction tools defined in BIRD-interact~\cite{birdinteract}. These tools allow the system agent to inspect the database schema, retrieve column semantics and external knowledge, probe database values, and execute SQL queries.
Table~\ref{tab:datasource_tools} summarizes the action space used in our experiments.
BIRD-Interact additionally associates predefined costs with these actions; we do not use these action-level costs in our experiments so that the system agent can freely explore the datasource and therefore omit them here.

\paragraph{Two-stage user simulation.}
The user simulator processes each clarification question in two stages. An encoder LLM first determines whether the question can be answered from the information available in the task specification and, when applicable, maps the question to one or more corresponding annotated ambiguities. If the question is answerable, a decoder LLM generates a natural-language response conditioned on the clarification question, the reference SQL, the matched annotated ambiguities, and their corresponding reference-SQL snippets. Questions unsupported by the task specification are treated as unanswerable rather than answered using inferred information.
Following the validated function-driven protocol of~\cite{birdinteract}, the encoder's ambiguity mapping is also used for groundedness evaluation: an annotated ambiguity is considered addressed when a clarification question is mapped to it.

\section{Implementation Details}
\label{app:method_details}

\paragraph{\model.} We elaborate our \model's execution pipeline in Algorithm~\ref{alg:poolask}. Specifically, \texttt{pool\_generate} initializes the pool with a ranked list of clarification questions, each associated with the ambiguous phrase it targets. \texttt{pool\_next} exposes the highest-priority pending question to the user and records the response; \texttt{pool\_view} returns the current pool state; \texttt{pool\_drop} explicitly removes an item that has become resolved or irrelevant; and \texttt{pool\_add} inserts a newly discovered clarification need. Each pool item is also assigned a unique identifier by the pool implementation upon insertion, which is exposed to the agent as part of the pool state.

\begin{algorithm}[h]
\caption{\model}\label{alg:poolask}
\begin{algorithmic}[1]
\Require ambiguous query $Q$, datasource $\mathcal{D}$, exploration tools $\mathcal{T}$

\Phase{Stage 1: Data Exploration}
\State $\mathcal{C} \gets \{Q\}$ \hfill $\triangleright$ interaction context
\While{agent deems exploration incomplete}
    \State $t \gets \textsc{SelectTool}(\mathcal{T}, \mathcal{C})$
    \State $o \gets \textsc{Execute}(t, \mathcal{D})$
    \State $\mathcal{C} \gets \mathcal{C} \cup \{o\}$
\EndWhile
\Statex

\Phase{Stage 2: Pool Initialization}
\State $\mathcal{P} \gets \textsc{GeneratePool}(Q, \mathcal{C})$
    \hfill $\triangleright$ ranked list of clarification questions
\Statex

\Phase{Stage 3: Iterative Clarification}
\While{$\mathcal{P} \neq \emptyset$}
    \State $a \gets \textsc{AgentDecide}(\mathcal{P}, \mathcal{C})$
    \If{$a = \texttt{pool\_view}$}
        \State $\mathcal{C} \gets \mathcal{C} \cup \{\mathcal{P}\}$
            \hfill $\triangleright$ review pool state
    \ElsIf{$a = \texttt{pool\_next}$}
        \State $q^* \gets \textsc{PopTop}(\mathcal{P})$
            \hfill $\triangleright$ highest-priority question
        \State $\mathit{ans} \gets \textsc{UserRespond}(q^*)$
        \State $\mathcal{C} \gets \mathcal{C} \cup \{(q^*,\, \mathit{ans})\}$
    \ElsIf{$a = \texttt{pool\_drop}(q_j)$}
        \State $\mathcal{P} \gets \mathcal{P} \setminus \{q_j\}$
            \hfill $\triangleright$ resolved by prior answer
    \ElsIf{$a = \texttt{pool\_add}(q')$}
        \State $\mathcal{P} \gets \mathcal{P} \cup \{q'\}$
            \hfill $\triangleright$ new ambiguity discovered
    \ElsIf{$a \in \mathcal{T}$}
        \State $\mathcal{C} \gets \mathcal{C} \cup \{\textsc{Execute}(a, \mathcal{D})\}$
            \hfill $\triangleright$ other exploration
    \EndIf
\EndWhile
\Statex

\Phase{Stage 4: SQL Generation}
\State \Return $\textsc{GenerateSQL}(Q, \mathcal{C})$
\end{algorithmic}
\end{algorithm}

\paragraph{Regenerate.} Unlike the prompt-only baselines, Regenerate requires a harness-level
intervention.
Upon the agent's submission attempt, the resulting SQL candidates are returned to the original agent through a divergence-audit observation.
Algorithm~\ref{alg:regenerate} summarizes the procedure.

\begin{algorithm}[h]
\caption{\textsc{Regenerate}: Divergence-Driven Ambiguity Discovery}\label{alg:regenerate}
\begin{algorithmic}[1]
\Require ambiguous query $Q$, datasource $\mathcal{D}$, exploration tools $\mathcal{T}$

\Phase{Stage 1: Exploration and Clarification (Branch 1)}
\State $\mathcal{C} \gets \{Q\}$
\While{agent deems exploration/clarification incomplete}
    \State $a \gets \textsc{AgentDecide}(\mathcal{C})$
    \If{$a \in \mathcal{T}$}
        \State $\mathcal{C} \gets \mathcal{C} \cup \{\textsc{Execute}(a, \mathcal{D})\}$
    \ElsIf{$a = \texttt{ask}(q)$}
        \State $\mathcal{C} \gets \mathcal{C} \cup \{(q,\, \textsc{UserRespond}(q))\}$
    \EndIf
\EndWhile
\State $\text{SQL}_1 \gets \textsc{GenerateSQL}(Q, \mathcal{C})$
    \hfill $\triangleright$ intercept first submit
\Statex

\Phase{Stage 2: Independent Regeneration (Branch 2)}
\State $\mathcal{C}_{\text{shared}} \gets \textsc{TruncateAtLastClarification}(\mathcal{C})$
\State $\text{SQL}_2 \gets \textsc{IndependentSession}(Q, \mathcal{C}_{\text{shared}}, \mathcal{D}, \mathcal{T})$
    \hfill $\triangleright$ full multi-turn session
\Statex

\Phase{Stage 3: Divergence Comparison}
\If{$\textsc{Normalize}(\text{SQL}_1) = \textsc{Normalize}(\text{SQL}_2)$}
    \State \Return $\text{SQL}_1$
        \hfill $\triangleright$ high confidence: pass through
\EndIf
\Statex

\Phase{Stage 4: Ambiguity Resolution (Branch 3)}
\State Present $\text{SQL}_1$, $\text{SQL}_2$ to agent in fresh context $\mathcal{C}_{\text{shared}}$
\While{agent identifies unresolved ambiguities from the divergence}
    \State $q \gets \textsc{FormulateQuestion}(\text{SQL}_1, \text{SQL}_2, \mathcal{C}_{\text{shared}})$
    \State $\mathit{ans} \gets \textsc{UserRespond}(q)$
    \State $\mathcal{C}_{\text{shared}} \gets \mathcal{C}_{\text{shared}} \cup \{(q,\, \mathit{ans})\}$
\EndWhile
\State \Return $\textsc{GenerateSQL}(Q, \mathcal{C}_{\text{shared}}, \text{SQL}_1, \text{SQL}_2)$
    \hfill $\triangleright$ generate final SQL
\end{algorithmic}
\end{algorithm}

\paragraph{Read-all.} Before the first agent turn, we execute the datasource inspection calls needed to retrieve the complete schema, column descriptions, and available external knowledge, and inject their outputs into the initial interaction history. The agent then proceeds with the same base harness, except that redundant exploration calls are discouraged by the additional instruction.

\section{Case Study}
\label{app:case_study}

We illustrate the behavior of \model using a representative example from BIRD-Interact-Lite (\texttt{museum\_3}), with the initial user query:
\begin{quote}
\emph{Calculate and rank all artifacts to identify which need urgent care.
The report must include Artifact ID, name, its historical era, and the score.}
\end{quote}
The instance contains three annotated ambiguities:
(1) what criterion defines urgent care,
(2) how the artifacts should be ranked, and
(3) how the domain-specific Conservation Priority Index (CPI) should be
computed, as the CPI definition cannot be determined from the database schema alone.

\paragraph{Initial clarification pool.}
After exploring the schema and available external knowledge, the agent identifies CPI as potentially relevant and initializes the following ranked pool:

\begin{enumerate}
    \item[\textbf{Q0.}] Which score should be used for ranking-the Artifact Vulnerability Score (AVS $=$ CPI $\times$ ERF), or another score?
    \item[\textbf{Q1.}] Which sensitivity columns should be averaged for the Environmental Risk Factor (ERF)?
    \item[\textbf{Q2.}] Should the \texttt{ConservePriorityScore} column be used directly as CPI?
    \item[\textbf{Q3.}] Does ``historical era'' correspond to \texttt{ArtDynasty}?
    \item[\textbf{Q4.}] Should artifacts be sorted in descending score order, and should an explicit numeric rank column be returned?
\end{enumerate}

Then, \model evolves as new information arrives, which is summarized in Table~\ref{tab:case_pool} below. For readability, Table~\ref{tab:case_pool} denotes \texttt{pool\_next()} as \texttt{pool\_next(Q\_i)} to indicate the pool item returned at that step.

\begin{table}[h]
\centering
\caption{Evolution of the clarification pool of \model in the case study.}
\label{tab:case_pool}
\small
\setlength{\tabcolsep}{4pt}
\begin{tabular}{p{0.05\linewidth} p{0.20\linewidth} p{0.7\linewidth}}
\toprule
\textbf{Step} & \textbf{Pool Action} & \textbf{Outcome} \\
\midrule

1 &
\texttt{pool\_next(Q0)} &
The user specifies that artifacts should be ranked directly by CPI rather
than AVS:
\textit{``Just use the Conservation Priority Index (CPI) itself---rank by CPI,
no AVS needed.''} \\

2 &
\texttt{pool\_drop(Q1)} &
As AVS is no longer needed, ERF becomes irrelevant.
The agent explicitly removes Q1 instead of asking an unnecessary question,
reasoning that CPI alone is sufficient. \\

3 &
\texttt{pool\_next(Q2)} &
The user clarifies that CPI should not be taken directly from
\texttt{ConservePriorityScore}:
\textit{``No, don't use that column. CPI needs to be calculated from the
artifact's historical significance, research value, cultural score, and
conservation status.''} \\

4 &
\texttt{pool\_add(Q5)} &
The previous answer reveals a new clarification need.
The agent adds the following question with highest priority:
\textit{``What is the exact CPI formula? How are the ratings combined, and
what is the numerical mapping for ConserveStatus?''} \\

5 &
\texttt{pool\_next(Q5)} &
The user provides the specific formula to calculate CPI. \\

6 &
\texttt{pool\_next(Q3)} &
The user confirms that historical era corresponds to
\texttt{ArtDynasty}. \\

7 &
\texttt{pool\_next(Q4)} &
The user confirms that artifacts should be sorted in descending order by CPI,
with no separate numeric rank column. \\
\bottomrule
\end{tabular}
\end{table}

\paragraph{Final SQL.}
After all pending pool entries have been answered or explicitly removed,
the agent submits the following SQL, which passes execution evaluation:

\begin{center}
\begin{minipage}{0.55\linewidth}
\begin{lstlisting}[style=promptstyle, basicstyle=\normalsize\ttfamily]
SELECT
    a.artregistry AS artifact_id,
    a.artname AS name,
    a.artdynasty AS historical_era,
    ((r.histsignrating
      + r.researchvalrating
      + r.cultscore)
     * (10 - CASE a.conservestatus
          WHEN 'Excellent' THEN 1
          WHEN 'Good'      THEN 3
          WHEN 'Fair'      THEN 5
          WHEN 'Poor'      THEN 7
          WHEN 'Critical'  THEN 10
        END)) / 30.0 AS cpi_score
FROM artifactscore a
JOIN artifactratings r
  ON a.artregistry = r.artref
ORDER BY cpi_score DESC;
\end{lstlisting}
\end{minipage}
\end{center}

\paragraph{Takeaway.}
This example highlights the two forms of adaptation enabled by \model. First, \texttt{pool\_drop} allows the agent to explicitly discharge a planned question that becomes irrelevant, avoiding rigid execution of the initial plan. Second, \texttt{pool\_add} allows newly revealed clarification needs to be incorporated into the same persistent state. The initial pool contains five questions; during interaction one question is dropped and one new question is added, resulting in five actual user questions. Thus, the pool provides persistent accounting of unresolved clarification needs without requiring the initial plan to remain fixed.


\begin{figure}[t]
\centering

\begin{tcolorbox}[
    colback=white,
    colframe=black,
    boxrule=0.7pt,
    arc=1mm,
    left=4pt,
    right=4pt,
    top=4pt,
    bottom=4pt,
    title={\textbf{Base System Prompt (1/2)}},
    coltitle=black,
    colbacktitle=white,
    fonttitle=\normalsize
]
\begin{lstlisting}[style=promptstyle]
You are a helpful {dialect} agent that interacts with a user and a database to solve the user's question.

# Task Description
Your goal is to understand the user's ambiguous question and generate the correct SQL query to solve it. You can:
1. Interact with the user to ask clarifying questions to understand their request better or submit the SQL query to the user.
2. Interact with the {dialect} environment to explore the database schema and data.
- Before submitting, interact with the environment and the user as much as the task requires: explore the schema, look up column meanings and external knowledge, and ask the user clarifying questions whenever the request is ambiguous.
- The task ends when you submit. You may submit only once and that submission is final.

# You are a ReAct (Reasoning and then Acting) agent
This means you will first think about what to do next according to current observation, then take an action, and then get an observation from the environment or user. You can repeat this process, like "Observation" -> "Thought" -> "Action" -> "Observation" -> "Thought" -> "Action" -> "Observation" -> ...

## Interaction Format (Response Format)
Given previous interaction history, and current observation (from the your previous interaction (env or user) or the user's request at the beginning), you should respond using the following format:

<thought> the agent's thought about the current state </thought> <interaction_object> interaction_object </interaction_object> <action> action </action>

## The interaction object and action space
- interaction_object: `Environment`
    - action: `execute(sql)` to interact with {language} database.
        - inputs:
            - sql: string, {language} command to execute. Could contain multiple commands separated by semicolon. MUST BE IN ONE STRING, ENCLOSED BY TWO QUOTES OR """YOUR SQL HERE""".
        - output: fetched result from {language} database.
    - action: `get_schema()` to get the schema of the database.
        - output: string of database schema in DDL format with demo data.
    - action: `get_all_column_meanings()` to get the meaning of all columns in the database.
        - output: string of all column meanings.
    - action: `get_column_meaning(table_name, column_name)` to get the meaning of a column.
        - inputs:
            - table_name: string, name of the table to get column meaning.
            - column_name: string, name of the column to get meaning.
        - output: string of column meaning.
    - action: `get_all_external_knowledge_names()` to get all external knowledge names.
        - output: list of string of external knowledge names.
    - action: `get_knowledge_definition(knowledge_name)` to get external knowledge by name.
        - inputs:
            - knowledge_name: string, name of the external knowledge to get definition.
        - output: string of external knowledge definition.
    - action: `get_all_knowledge_definitions()` to get all external knowledge names with definitions.
        - output: string of all external knowledge names with definitions.
    - action: `probe_value(table_name, column_name, value)` to check if a specific value exists in a column. If no exact match, returns similar values.
        - inputs:
            - table_name: string, name of the table.
            - column_name: string, name of the column.
            - value: string, the value to search for.
        - output: exact_match (true/false) and a list of similar_values.

- interaction_object: `User`
    - action: `ask(question)` to ask user for clarification. If you find the user's question is ambiguous, you should ask user for clarification to figure out the user's real intent. YOU ARE ONLY ALLOWED TO ASK ONE QUESTION AT A TIME.
        - inputs:
            - question: string, question to ask user for clarification.
        - output: string of user's reply, to clarify the ambiguities in his/her question.
    - action: `submit(sql)` to submit the SQL to the user. This is your FINAL answer and ends the task.
        - inputs:
            - sql: string, SQL to submit to the user. Could contain multiple commands separated by semicolon. MUST BE IN ONE STRING, ENCLOSED BY TWO QUOTES OR """YOUR SQL HERE""".
        - output: feedback from user about the submitted SQL.
\end{lstlisting} 
\end{tcolorbox}
\vspace{-0.3cm}
\caption{Base system prompt shared by all agentic methods (Part 1/2).}
\label{fig:base_prompt_1}
\end{figure}

\begin{figure}[t]
\centering

\begin{tcolorbox}[
    colback=white,
    colframe=black,
    boxrule=0.7pt,
    arc=1mm,
    left=4pt,
    right=4pt,
    top=4pt,
    bottom=4pt,
    title={\textbf{Base System Prompt (2/2)}},
    coltitle=black,
    colbacktitle=white,
    fonttitle=\normalsize
]
\begin{lstlisting}[style=promptstyle]
# Important Strategy Tips
- First explore the database schema, column meaning and external knowledge to understand available tables, columns and user query's involved external knowledge.
- FIGURE OUT THE USER'S REAL INTENT BY ASKING CLARIFYING QUESTIONS! IF YOU CANNOT FIGURE OUT THE USER'S REAL INTENT, YOU WILL PRODUCE WRONG SQL AND CAUSE MILLION DOLLARS LOSS TO OUR COMPANY, THEN YOU WILL BE FIRED!!! (YOU ARE ONLY ALLOWED TO ASK ONE QUESTION AT A TIME.)
- FIGURE OUT THE USER'S REAL INTENT BY ASKING CLARIFYING QUESTIONS! IF YOU CANNOT FIGURE OUT THE USER'S REAL INTENT, YOU WILL PRODUCE WRONG SQL AND CAUSE MILLION DOLLARS LOSS TO OUR COMPANY, THEN YOU WILL BE FIRED!!! (YOU ARE ONLY ALLOWED TO ASK ONE QUESTION AT A TIME.)
- Make sure your submitted SQL is valid and addresses all aspects of the question.
- Resolve every ambiguity and verify your understanding before you submit, since the submission is final.

+ query_msg: "# -----TASK START-----\nNow, let's start with the user's question that may exist ambiguities and require external knowledge understanding to solve. (EACH TIME GIVE ONE ROUND RESPONSE, END YOUR RESPONSE AT ... '</action>' OTHERWISE YOU WILL BE FIRED!!!)\n\nUser's Question: {query}\n:"
\end{lstlisting}
\end{tcolorbox}

\caption{Base system prompt shared by all agentic methods (Part 2/2).}
\label{fig:base_prompt_2}
\end{figure}

\begin{figure}[t]
\centering
\begin{tcolorbox}[
    colback=white,
    colframe=black,
    boxrule=0.7pt,
    arc=1mm,
    left=4pt,
    right=4pt,
    top=4pt,
    bottom=4pt,
    title={\textbf{Read-all-specific system prompt}},
    coltitle=black,
    colbacktitle=white,
    fonttitle=\small
]
\begin{lstlisting}[style=promptstyle]
[SYSTEM NOTE: The database's COMPLETE information has already been retrieved for you and is shown in the interaction history below. You do NOT need to call get_schema(), get_all_column_meanings(), get_all_knowledge_definitions(), or any other get_* exploration action; all of it is already available to you. If the user's request itself is still ambiguous, you may ask the user clarifying questions (one ask(...) per turn). When you are confident you understand the request, submit your final SQL.]
\end{lstlisting}
\end{tcolorbox}
\caption{Additional system instruction used by the Read-All baseline.}
\label{fig:prompt_readall}
\end{figure}

\begin{figure}[t]
\centering
\begin{tcolorbox}[
    colback=white,
    colframe=black,
    boxrule=0.7pt,
    arc=1mm,
    left=4pt,
    right=4pt,
    top=4pt,
    bottom=4pt,
    title={\textbf{Regenerate-specific system prompt}},
    coltitle=black,
    colbacktitle=white,
    fonttitle=\small
]
\begin{lstlisting}[style=promptstyle]
Observation: Two SQL implementations were generated independently from your interaction context. They differ in some implementation decisions:

Candidate A:
{sql1}

Candidate B:
{sql2}

Analyze whether the differences between them reveal unresolved ambiguities in the user's request, or are merely equivalent implementation. If the differences are purely equivalent, submit either candidate as-is. If they reveal genuine ambiguities you missed, ask the user to clarify each one, then revise and submit.
\end{lstlisting}
\end{tcolorbox}
\caption{Runtime observation injected by the Regenerate baseline after two independently generated SQL candidates diverge.}
\label{fig:prompt_regenerate}
\end{figure}

\begin{figure}[t]
\centering
\begin{tcolorbox}[
    colback=white,
    colframe=black,
    boxrule=0.7pt,
    arc=1mm,
    left=4pt,
    right=4pt,
    top=4pt,
    bottom=4pt,
    title={\textbf{Draft-specific system prompt}},
    coltitle=black,
    colbacktitle=white,
    fonttitle=\small
]
\begin{lstlisting}[style=promptstyle]
[SYSTEM NOTE: Follow a DRAFT-THEN-ASK strategy.
1. First, explore the database as much as you need (schema, column meanings, external knowledge).
2. When you have enough context, write a COMPLETE best-effort draft SQL in your thought. Make your best guess for every ambiguous decision.
3. Then AUDIT your draft: identify every implementation decision you made WITHOUT explicit support from the user's request, schema, or knowledge base.
   For each low-confidence decision, formulate a targeted clarification question.
   Your questions should be SPECIFIC to your draft, e.g.:
   - "I'm computing X as [formula]. Is this correct, or should it be [alternative]?"
   - "I'm outputting column Y as [type/format]. Do you want [alternative] instead?"
   - "I'm using [JOIN type] to connect A and B. Should it be [other JOIN] instead?"
   - "For rows where [condition] is unknown, I default to [value]. Is that right?"
4. Ask your questions ONE AT A TIME using ask(...), most important first.
5. After all questions are answered, REVISE your draft based on the answers and submit your final SQL.]
\end{lstlisting}
\end{tcolorbox}
\caption{Additional system instruction used by the Draft baseline.}
\label{fig:prompt_draft}
\end{figure}

\begin{figure}[t]
\centering
\begin{tcolorbox}[
    colback=white,
    colframe=black,
    boxrule=0.7pt,
    arc=1mm,
    left=4pt,
    right=4pt,
    top=4pt,
    bottom=4pt,
    title={\textbf{Self-reflection-specific system prompt}},
    coltitle=black,
    colbacktitle=white,
    fonttitle=\small
]
\begin{lstlisting}[style=promptstyle]
Your SQL has been drafted but NOT yet submitted.

Your draft:
{sql}

Before final submission, review your draft:
- Identify any implementation choice where the user's answer or the knowledge base did not explicitly specify what to do, and you had to make an assumption.
- For each such assumption, decide: is there a reasonable alternative that would produce different results? If yes, ask the user to confirm.

{ask_instruction}
You may use ask(question) to clarify with the user, or submit(SQL) to finalize.
\end{lstlisting}
\end{tcolorbox}
\caption{Additional system instruction used by the Self-Reflection baseline.}
\label{fig:prompt_reflection}
\end{figure}

\begin{figure}[t]
\centering
\begin{tcolorbox}[
    colback=white,
    colframe=black,
    boxrule=0.7pt,
    arc=1mm,
    left=4pt,
    right=4pt,
    top=4pt,
    bottom=4pt,
    title={\textbf{Plan-Prompt-specific system prompt}},
    coltitle=black,
    colbacktitle=white,
    fonttitle=\small
]
\begin{lstlisting}[style=promptstyle]
[SYSTEM NOTE: Follow a PLAN-THEN-ASK strategy.
1. First, explore the database as much as you need to understand the request. Do NOT ask the user anything yet.
2. Then, in a SINGLE thought, write a complete, ranked list of EVERY ambiguity in the request that you need the user to clarify. Begin that thought with a line 'PLAN:' followed by a numbered list, most important first.
3. Then ask your planned questions ONE AT A TIME (one ask(...) per turn), in your ranked order.
4. After your planned questions are answered, submit your final SQL.]
\end{lstlisting}
\end{tcolorbox}
\caption{Additional system instruction used by the Plan-Prompt baseline.}
\label{fig:prompt_planprompt}
\end{figure}

\begin{figure}[t]
\centering
\begin{tcolorbox}[
    colback=white,
    colframe=black,
    boxrule=0.7pt,
    arc=1mm,
    left=4pt,
    right=4pt,
    top=4pt,
    bottom=4pt,
    title={\textbf{\model-specific system prompt}},
    coltitle=black,
    colbacktitle=white,
    fonttitle=\small
]
\begin{lstlisting}[style=promptstyle]
[SYSTEM NOTE: Follow a POOL-ASK strategy.
1. First, explore the database as much as you need to understand the request.
2. When ready, call pool_generate(<json_array>) with a JSON array of ALL ambiguities you want to clarify. Format:
   [{"question": "...", "phrase": "...", "rank": 1}, ...]
   Fields:
   - "question": the exact clarification question for the user
   - "phrase": the ambiguous phrase/term from the request this resolves
   - "rank": priority (1 = most important, largest impact on SQL)
3. After pool_generate, call pool_next() each time you are ready to receive the next answer.
4. Between pool_next() calls, you may use these OPTIONAL pool management tools:
   - pool_view()              -- see the full pool status (all items + answers)
   - pool_drop(id, "reason")  -- drop a pending item that is now unnecessary
   - pool_add({"question": "...", "phrase": "...", "rank": N})  -- add a new item
5. When all pool items are answered or dropped, submit your final SQL.]
\end{lstlisting}
\end{tcolorbox}
\caption{Additional system instruction used by \model.}
\label{fig:prompt_planpool}
\end{figure}

\end{document}